\documentclass[10pt]{article}
\usepackage{graphicx}
\usepackage{natbib}
\usepackage{url} 
\usepackage{amssymb,amsmath,amsthm,graphicx,bbm}
\usepackage[linesnumbered,ruled]{algorithm2e}
\usepackage{algorithmic}
\usepackage[toc,page]{appendix}
\usepackage{rotating}

\newcommand{\blind}{0}

\usepackage{amssymb}
\begin{document}

\def\spacingset#1{\renewcommand{\baselinestretch}%
{#1}\small\normalsize} \spacingset{1}

\if0\blind
{
  \title{\bf A Comparison of Machine Learning Methods for Data with High-Cardinality Categorical Variables}
  
  \author{Fabio Sigrist\thanks{Email: fabio.sigrist@hslu.ch. Address: Lucerne University of Applied Sciences and Arts, Suurstoffi 1, 6343 Rotkreuz, Switzerland.}\\
   Lucerne University of Applied Sciences and Arts}
  \maketitle
}
\fi

\bigskip

\spacingset{1} 

\providecommand{\keywords}[1]
{
	\small
	\textbf{\textit{Keywords:}} #1
	\normalsize
}

\begin{abstract}%
	High-cardinality categorical variables are variables for which the number of different levels is large relative to the sample size of a data set, or in other words, there are few data points per level. Machine learning methods can have difficulties with high-cardinality variables. In this article, we empirically compare several versions of two of the most successful machine learning methods, tree-boosting and deep neural networks, and linear mixed effects models using multiple tabular data sets with high-cardinality categorical variables. We find that, first, machine learning models with random effects have higher prediction accuracy than their classical counterparts without random effects, and, second, tree-boosting with random effects outperforms deep neural networks with random effects.
\end{abstract}


\section{Introduction}\label{intro}

High-cardinality categorical variables are variables for which the number of different levels is large relative to the sample size of a data set or, equivalently, there is little data per level. High-cardinality categorical variables can pose difficulties for machine learning methods such as deep neural networks and tree-based models. 

A simple strategy for dealing with categorical variables is to use one-hot encoding or dummy variables. But this approach often does not work well for high-cardinality categorical variables due to the reasons described below. For neural networks, a frequently adopted solution is to use entity embeddings \citep{guo2016entity} that map every level of a categorical variable into a low-dimensional Euclidean space. For tree-boosting, an alternative to one-hot encoding is to assign a number to every level of a categorical variable, and then consider this as a one-dimensional numeric variable. Another solution implemented in the \texttt{LightGBM} boosting library \citep{ke2017lightgbm} works by partitioning all levels into two subsets using an approximate approach \citep{fisher1958grouping} when finding splits in the tree-building algorithm. Further, the \texttt{CatBoost} boosting library \citep{CatBoost2017} implements an approach based on ordered target statistics calculated using random partitions of the training data for handling categorical predictor variables.

Random effects \citep{laird1982random, pinheiro2006mixed} can also be used as a tool for handling high-cardinality categorical variables. In a random effects model, it is assumed that a (potentially transformed) parameter $\mu\in\mathbb{R}^{n}$ of the response variable distribution equals the sum of fixed $F(X)$ and random effects $Zb$:
\begin{equation*}
	\mu=F(X)+Zb,~~~~b\sim \mathcal{N}(0,\Sigma),
\end{equation*}
where $F(X)$ is the row-wise evaluation of a function $F(\cdot):\mathbb{R}^p\rightarrow\mathbb{R}$, $F(X)=(F(X_1),\dots,F(X_n))^T$, $X_{i}=(X_{i1}\dots,X_{ip})^T\in\mathbb{R}^{p}$ is the $i$-th row of the fixed effects variables matrix $X\in\mathbb{R}^{n\times p}$, $i=1,\dots,n$,  $b \in \mathbb{R}^m$, and $Z\in\mathbb{R}^{n\times m}$. These models are called mixed effects models since they contain both fixed effects $F(X)$ and random effects $Zb$. If the conditional response variable distribution is Gaussian and there is a single high-cardinality categorical variable, such a mixed effects model can also be written as
\begin{equation}\label{ex_re_model}
	y_{ij} = F(x_{ij}) + b_i + \epsilon_{ij},~~  b_i \overset{iid}{\sim } \mathcal{N}(0,\sigma^2_1), ~~ \epsilon_{ij} \overset{iid}{\sim } \mathcal{N}(0,\sigma^2),
\end{equation}
where $j=1,\dots,n_i$ is the sample index within level $i$ with $n_i$ being the number of samples for which the categorical variable attains level $i$, $i=1,\dots,q$ is the level index with $q$ being the number of levels of the categorical variable, and $x_{ij}$ are the fixed effects predictor variables for observation $ij$. The total number of samples is $n=\sum_{i=1}^qn_i$. Further, the random effects $b_i$ and $\epsilon_{ij}$ are assumed to be independent. For this model, the matrix $Z$ is simply a binary incidence matrix that maps every random effect $b_i$ to its corresponding observations and $\Sigma = \sigma^2 I_m$.

In (generalized) linear mixed effects models it is assumed that $F(\cdot)$ is a linear function: $F(X)=X\beta$. In the last years, linear mixed effects models have been extended to non-linear ones using single trees \citep{hajjem2011mixed,sela2012re, fu2015unbiased}, random forest \citep{hajjem2014mixed}, tree-boosting \citep{sigrist2022gaussian, sigrist2023gaussian}, and most recently (in terms of first public preprint) deep neural networks \citep{simchoni2021using, simchoni2023integrating, avanzi2023machine}. In contrast to classical independent machine learning models, the random effects introduce dependence among samples.

\subsection{Why are random effects useful for high-cardinality categorical variables?}
For high-cardinality categorical variables, there is little data for every level. Intuitively, if the response variable has a different (conditional) mean for many levels, traditional machine learning models (with, e.g., one-hot encoding, embeddings, or simply one-dimensional numeric variables) may have problems with over- or underfitting for such data. From the point of view of a classical bias-variance trade-off, independent machine learning models may have difficulties balancing this trade-off and finding an appropriate amount of regularization. For instance, overfitting may occur which means that a model has a low bias but high variance. 

Broadly speaking, random effects act as a prior, or regularizer, which models the difficult part of a function, i.e., the part whose ``dimension" is similar to the total sample size, and, in doing so, provide an effective way for finding a balance between over- and underfitting or bias and variance. For instance, for a single categorical variable, random effects models will shrink estimates of group intercept effects towards the global mean. This process is sometimes also called ``information pooling". It represents a trade-off between completely ignoring the categorical variable (= underfitting / high bias and low variance) and giving every level in the categorical variable ``complete freedom" in estimation (= overfitting / low bias and high variance). Importantly, the amount of regularization, which is determined by the variance parameters of the model, is learned from the data. Specifically, in the above single-level random effects model in \eqref{ex_re_model}, a (point) prediction $\hat y^p$ for the response variable for a sample with predictor variables $x^p$ and categorical variable having level $i$ is given by 
\begin{equation*}
	\hat y^p = \hat F(x^p) + \frac{\hat\sigma^2_1}{\hat\sigma^2/n_i + \hat\sigma^2_1}(\bar y_i - \bar F_i),
\end{equation*}
where $\hat F(x^p)$ is the trained function evaluated at $x^p$, $\hat\sigma^2_1$ and $\hat\sigma^2$ are variance estimates, and $\bar y_i$ and $\bar F_i$ are sample means of $y_{ij}$ and $F(x_{ij})$, respectively, for level $i$. Ignoring the categorical variable would give the prediction $\hat y^p = \hat F(x^p)$, and a fully flexible model without regularization gives $\hat y^p = \hat F(x^p)  + (\bar y_i - \bar F_i)$. I.e., the difference between these two extreme cases and the random effects model is the shrinkage factor $\frac{\hat\sigma^2_1}{\hat\sigma^2/n_i + \hat\sigma^2_1}$ (which goes to zero if the number of samples $n_i$ for level $i$ is large). Related to this, random effects models allow for more efficient (i.e., lower variance) estimation of the fixed effects function $F(\cdot)$ \citep{sigrist2022gaussian}. See also \citet[Section 1.1]{sigrist2023gaussian} for a discussion on why random effects are useful for modeling high-cardinality categorical variables. 

In line with the above argumentation, \citet[Section 4.1]{sigrist2023gaussian} find in empirical experiments that tree-boosting combined with random effects outperforms traditional independent tree-boosting the more, the lower the number of samples per level of a categorical variable, i.e., the higher the cardinality of a categorical variable.

\section{Methods, data sets, and experimental settings}

In the following, we compare several methods using multiple real-world data sets with high-cardinality categorical variables. We use all the publicly available tabular data sets from \citet{simchoni2021using, simchoni2023integrating} and also the same experimental setting as in \citet{simchoni2021using, simchoni2023integrating}.  In addition, we include the Wages data set analyzed in \citet{sigrist2022gaussian}.

\subsection{Methods}

We consider the following methods: 
\begin{itemize}
	\item `Linear': linear mixed effects models
	\item `NN\_Embed': deep neural networks with embeddings
	\item `LMMNN': combining deep neural networks and random effects \citep{simchoni2021using, simchoni2023integrating}
	\item `LGBM\_Num': tree-boosting by assigning a number to every level of categorical variables and considering these as one-dimensional numeric variables
	\item `LGBM\_Cat': tree-boosting with the approach of \texttt{LightGBM} \citep{ke2017lightgbm} for categorical variables
	\item `CatBoost': tree-boosting with the approach of \texttt{CatBoost} \citep{CatBoost2017} for categorical variables
	\item `GPBoost': combining tree-boosting and random effects \citep{sigrist2022gaussian, sigrist2023gaussian}
\end{itemize}
The MERF algorithm of \citet{hajjem2014mixed} is another promising mixed effects machine learning method, but its current implementation in the form of a Python package\footnote{https://github.com/manifoldai/merf} is prohibitively slow for the sample sizes of the data sets considered here. Also note that, recently (starting with version 1.6), the \texttt{XGBoost} library \citep{chen2016xgboost} has also implemented the same approach as \texttt{LightGBM} for handling categorical variables.\footnote{See https://xgboost.readthedocs.io/en/stable/tutorials/categorical.html\#optimal-partitioning (retrieved on June 30, 2023)} We do not consider this as a separate approach here.

\subsection{Data sets}
Table \ref{desc_data} gives an overview of the data sets. For more details on the data, we refer to \citet{simchoni2021using, simchoni2023integrating} and \citet{sigrist2022gaussian}.

\begin{table}[ht!]
	\centering
	\begingroup
	\begin{tabular}{lllllll}
		\hline
		\hline
		Dataset & $n$ & $p$ & $K$ & Cat. var. & Nb. levels & Response var.\\ 
		\hline
		Airbnb & $50$K & $196$ & $1$ & host & $39$K & price (log)  \\
		\hline
		IMDb & $88$K & $159$ & $2$ & director & $38$K & avg. movie score \\ 
		&  & & & movie type & $1.7$K &  \\ 
		\hline
		Spotify & $28$K & $14$ & $4$ & artist & $10$K & song danceability\\
		&  & & & album & $22$K &  \\ 
		&  & & & playlist & $2.3$K &  \\ 
		&  & & & subgenre & $553$ &  \\ 
		\hline
		News & $81$K & $176$ & $2$ & source & $5.4$K  &  nb. shares (log) \\ 
		&  & & & title & $72$K &  \\ 
		\hline
		InstEval & $73$K & $3$ & $3$ & student  & $2.9$K & teacher rating  \\ 
		&  & & & teacher & $1.1$K &  \\ 
		&  & & & department & $14$ &  \\ 
		\hline
		Rossmann & $33$K & $23$ & $1$ & store  & $1.1$K & total sales (in 100K) \\   
		\hline
		AUimport & $125$K & $8$ & $1$ & commodity  & $5$K & total import (log) \\  
		\hline
		Wages & $28$K & $52$ & $1$ & person  & $4.7$K & wage (log)\\ 
		\hline
		\hline
	\end{tabular}
	\endgroup
	\caption{Summary of data sets. $n$ is the number of samples, $p$ is the number of predictor variables (excl. high-cardinality categorical variables), $K$ is the number of high-cardinality categorical variables, and 'Cat. var.' describes the categorical variable(s). } 
	\label{desc_data}
\end{table}

For all methods with random effects, we include random effects for every categorical variable mentioned in Table \ref{desc_data} with no prior correlation among random effects. The Rossmann, AUImport, and Wages data sets are longitudinal data sets. This means that the samples for every level or the categorical variables are repeated measurements over time $t$. As in \citet{simchoni2023integrating}, we additionally include random coefficients (= random slopes), for the time variables $t$ and $t^2$ besides intercept random effects with no prior correlation among the random effects. I.e., these random effects models are given by
\begin{equation*}
	y_{ij} = F(x_{ij}) + b_{0,i} + b_{1,i} \cdot t_{ij}+ b_{2,i} \cdot t_{ij}^2+ \epsilon_{ij},~~  b_{k,i} \overset{iid}{\sim } \mathcal{N}(0,\sigma^2_k), k\in\{1,2,3\}, ~~ \epsilon_{ij} \overset{iid}{\sim } \mathcal{N}(0,\sigma^2),
\end{equation*}
where $b_{1,i}$ and $b_{2,i}$ are the random slopes.

\subsection{Experimental setting}

We use the same experimental setting as in \citet{simchoni2023integrating} to compare the different methods. This means that we perform 5-fold cross-validation with the test mean squared error (MSE) to measure prediction accuracy. For the longitudinal data sets, we use the ``random mode" 5-fold CV setting of \citet{simchoni2023integrating} where in every fold, 80\% of the data is used for training and 20\% of the data for validation irrespective of the time variable $t$. The results for neural networks with embeddings and also neural networks with random effects (LMMNN) are taken from \citet{simchoni2021using, simchoni2023integrating}; see \citet{simchoni2021using, simchoni2023integrating} for more details on the specifications of these two modeling approaches. For linear mixed effects models, LGBM\_Num, LGBM\_Cat, GPBoost, and GPBoost\_I, we use the \texttt{GPBoost} library version 1.2.1 which is built upon the \texttt{LightGBM} library. Further, we use version 1.1.1 of the \texttt{CatBoost} library. 

For all tree-boosting methods, we choose tuning parameters on every of the five training sets in the $5$-fold CV by randomly splitting the training data into inner training data containing 80\% of the outer training data and validation data consisting of the remaining 20\%. We use a deterministic grid search with the mean squared error as a selection criterion and consider the following parameter combinations: number of boosting iterations $M\in \{1,\dots,1000\}$, learning rate $\in \{1,0.1,0.01\}$, maximal tree-depth $\in \{1,2,3,5,10\}$, minimal number of samples per leaf $\in \{10,100,1000\}$, and L2 penalty on leaf values $\in \{0,1,10\}$. For the machine learning models with random effects, one can either exclude or include the high-cardinality categorical variables in the fixed effects function. When additionally including them, one allows for potential interaction between the categorical variables and other predictor variables. For GPBoost, we consider this as a tuning parameter option. However, the difference between these two modeling options is minor which is an indication that there is no interaction present; see Table \ref{results_high_card_excl_incl_sep} in the appendix, where we report the results when separately either excluding or including the high-cardinality categorical variables in the fixed effects tree ensemble function.

Code for pre-processing the data with instructions on how to download the data and code for running the experiments can be found at \url{https://github.com/fabsig/Compare_ML_HighCardinality_Categorical_Variables}. Pre-processed data for modeling can also be found on the above webpage for data sets for which the license of the original source permits it.

\section{Results}

\begin{figure}[ht!]
	\centering
	\includegraphics[width=0.8\textwidth]{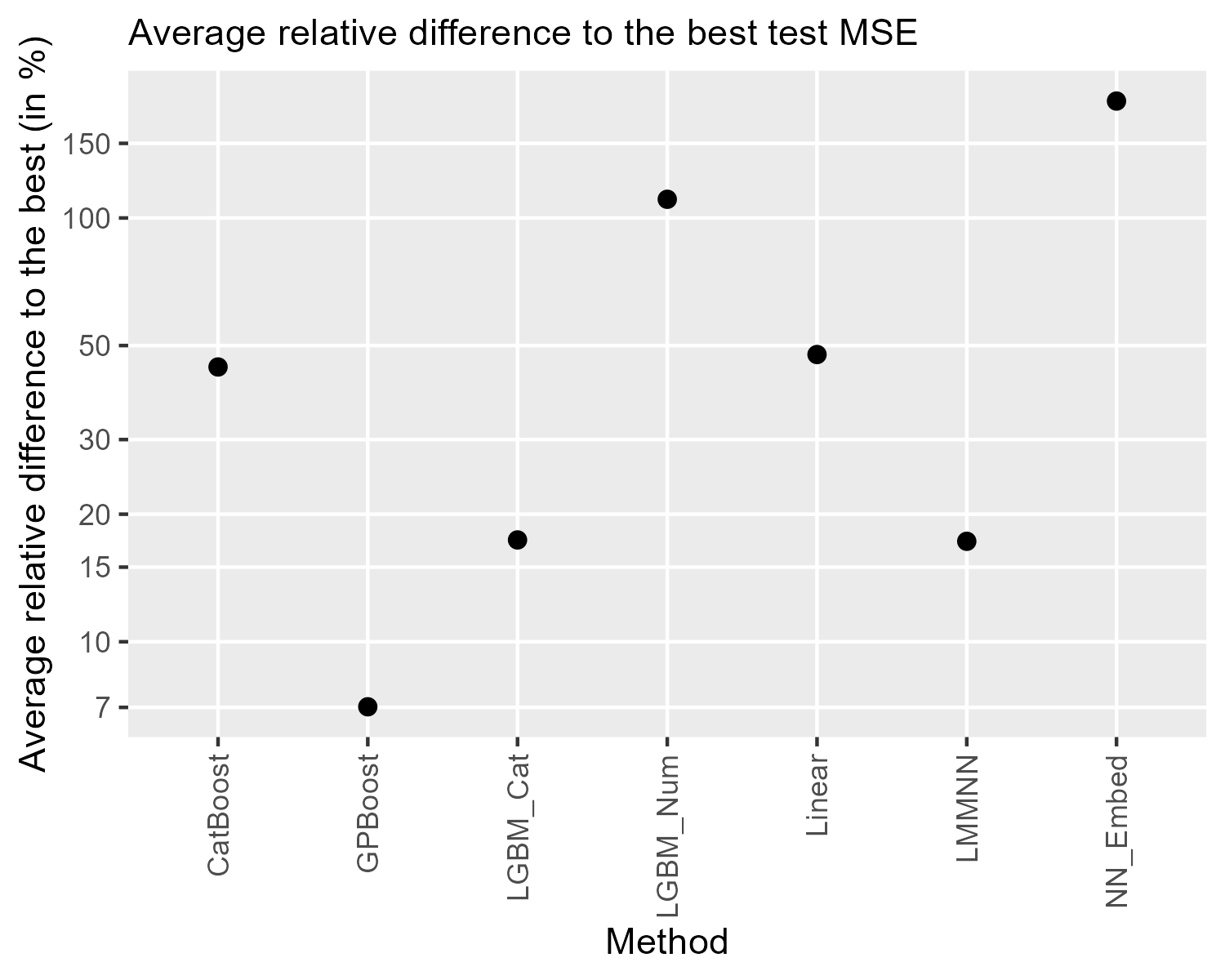}
	\caption{Average relative difference (in \%) to the lowest test MSE. The Wages data set is not included for calculating this since not all methods were run on it.}
	\label{fig_avg_rel_dif}
\end{figure}

The results are reported in Figure \ref{fig_avg_rel_dif} and Table \ref{results_high_card}. Table \ref{results_high_card} reports test mean squared errors (MSE) and corresponding standard errors. To summarize the prediction accuracy over the different data sets, we report average relative differences to the best result and average ranks. The former is obtained by calculating the relative difference of a test MSE of a method compared to the lowest MSE for every data set and then taking the average over all data sets. In Figure \ref{fig_avg_rel_dif}, we also report the average relative difference to the best result. In addition, Table \ref{time_high_card} in the appendix reports average wall clock times.

\begin{table}[ht!]
\centering
\begingroup\footnotesize
\begin{tabular}{llllllll}
  \hline
\hline
Dataset & Linear & LGBM\_Cat & LGBM\_Num & CatBoost & GPBoost & NN\_Embed & LMMNN \\ 
  \hline
Airbnb & 0.268 & 0.131 & 0.132 & 0.131 & \textbf{0.125} & 0.158 & 0.142 \\ 
   & (0.0748) & (0.00283) & (0.00285) & (0.00267) & (0.00264) & (0.00158) & (0.00167) \\ 
  IMDb & 1.02 & 0.954 & 1.05 & 0.922 & \textbf{0.850} & 1.26 & 0.974 \\ 
   & (0.00730) & (0.00721) & (0.00830) & (0.0117) & (0.00825) & (0.124) & (0.00933) \\ 
  Spotify & 0.0111 & 0.00858 & 0.00880 & \textbf{0.00851} & 0.00910 & 0.0164 & 0.00926 \\ 
   & (9.20e-05) & (6.23e-05) & (5.79e-05) & (4.69e-05) & (0.000164) & (0.000540) & (8.34e-05) \\ 
  News & 1.88 & 1.89 & 2.37 & 1.85 & \textbf{1.72} & 1.89 & 1.81 \\ 
   & (0.0146) & (0.0147) & (0.0157) & (0.0160) & (0.0131) & (0.0215) & (0.0190) \\ 
  InstEval & 1.44 & \textbf{1.44} & 1.53 & 1.46 & 1.47 & 1.50 & 1.45 \\ 
   & (0.00672) & (0.00746) & (0.00549) & (0.00711) & (0.0230) & (0.00669) & (0.00492) \\ 
  Rossmann & 0.0153 & \textbf{0.00627} & 0.0124 & 0.01000 & 0.00877 & 0.0516 & 0.0105 \\ 
   & (0.000552) & (0.000188) & (0.000848) & (0.000267) & (0.000284) & (0.00600) & (0.000162) \\ 
  AUimport & 0.743 & 1.26 & 4.56 & 2.15 & \textbf{0.651} & 3.35 & 0.713 \\ 
   & (0.0138) & (0.0269) & (0.0177) & (0.0477) & (0.00396) & (0.455) & (0.00700) \\ 
  Wages & 0.321 & 0.102 & 0.0992 & 0.0920 & \textbf{0.0830} &  &  \\ 
   & (0.0222) & (0.00259) & (0.00234) & (0.00251) & (0.00200) &  &  \\ 
   \hline
Avg. rank & 4.71 & 2.71 & 5.57 & 2.86 & \textbf{2.14} & 6.43 & 3.57 \\ 
  Avg. rel. dif. & 47.6 & 17.4 &    111. & 44.5 & \textbf{7.02} &    189. & 17.3 \\ 
   \hline
\hline
\end{tabular}
\endgroup
\caption{Average test mean squared error (MSE) and standard errors in parentheses. 
                  Best results are bold. 
                  'Avg. rank' denotes the average rank of a method.
                  'Avg. rel. dif.' denotes the average relative difference in \% of a method compared to the best method. 
                  The Wages data set is not included for calculating ranks and relative differences 
                  since not all methods were run on it.} 
\label{results_high_card}
\end{table}

The results in Table \ref{results_high_card} show that combined tree-boosting and random effects (GPBoost) has the highest prediction accuracy. GPBoost has an average relative difference to the best result of $7.02\%$ and an average rank of $2.14$. Combined neural networks and random effects (LMMNN) have an average relative difference to the best result of $17.3\%$ and an average rank of $3.57$. \texttt{LightGBM} (LGBM\_Cat) has a similar average relative difference to the best result of $17.4\%$ and an average rank of $2.71$. Next, CatBoost has an average relative difference to the best method of $44.5\%$ and an average rank of $2.86$. Linear mixed effects models perform worse having an average relative difference of approximately $47.6\%$ and an average rank of $4.71$. Overall worst perform neural networks with embeddings having an average relative difference to the best result of $189\%$. Tree-boosting with the categorical variables transformed to one-dimensional numeric variables (LGBM\_Num) performs slightly better with an average relative difference to the best result of $111\%$.\footnote{In their online documentation, \texttt{LightGBM} recommends ``For a categorical feature with high cardinality, it often works best to treat the feature as numeric ..."; see https://lightgbm.readthedocs.io/en/latest/Advanced-Topics.html\#categorical-feature-support (retrieved on June 30, 2023). We clearly come to a different conclusion. \nopagebreak}

\section{Conclusion}
We empirically compare various versions of tree-boosting and deep neural networks as well as linear mixed effects models on multiple tabular data sets with high-cardinality variables. We find that, first, machine learning models with random effects have higher prediction accuracy than their classical independent counterparts without random effects and, second, tree-boosting with random effects performs better than deep neural networks with random effects. While there may be several reasons for the latter finding, this is in line with the recent work of \citet{Grinsztajn2022} who find that tree-boosting outperforms deep neural networks (and also random forest) on tabular data without high-cardinality categorical variables. Similarly, \citet{shwartz2022tabular} conclude that tree-boosting ``outperforms deep models on tabular data."

\clearpage
\begin{appendices}
	
	\section{Additional results}\label{runtimes}
	
\begin{table}[ht!]
\centering
\begingroup\footnotesize
\begin{tabular}{llllllll}
  \hline
\hline
Dataset & Linear & LGBM\_Cat & LGBM\_Num & CatBoost & GPBoost & NN\_Embed & LMMNN \\ 
  \hline
Airbnb & 1.93 & 6.41 & 0.952 & 16.8 & 14.4 & 825. & 95.4 \\ 
  IMDb &    183. & 1.82 & 1.63 & 19.8 & 236. & 55.8 & 144. \\ 
  Spotify & 99.1 & 5.06 & 1.10 & 7.07 & 90.6 & 12.7 & 57.1 \\ 
  News & 138. & 1.02 & 1.54 & 93.1 & 63.9 & 31.1 & 83.9 \\ 
  InstEval & 95.3 & 4.64 & 0.773 & 25.5 & 682. & 51.1 & 92.5 \\ 
  Rossmann & 8.22 & 3.61 & 1.32 & 54.6 & 1.45 & 53.6 & 76.3 \\ 
  AUimport & 37.7 & 8.87 & 0.604 & 17.7 & 45.9 & 107. &    175. \\ 
  Wages & 19.6 & 3.61 & 2.41 & 43.6 & 34.2 &  &  \\ 
   \hline
\hline
\end{tabular}
\endgroup
\caption{Average wall-clock time in seconds. 
                  Note that except for the linear model, 
                  the wall-clock time depends on the chosen tuning parameters. The experiments for 
                  'NN\_Embed' and 'LMMNN' were run on Google Colab with NVIDIA Tesla V100 GPU machines 
                  (see Simchoni and Rosset, 2023), and all for all other methods, a laptop with an 
                  Intel i7-12800H processor was used.} 
\label{time_high_card}
\end{table}

\begin{table}[ht!]
\centering
\begingroup\footnotesize
\begin{tabular}{lllllllll}
  \hline
\hline
Dataset & Linear & LGBM\_Cat & LGBM\_Num & CatBoost & GPBoost\_E & GPBoost\_I & NN\_Embed & LMMNN \\ 
  \hline
Airbnb & 0.268 & 0.131 & 0.132 & 0.131 & \textbf{0.124} & 0.125 & 0.158 & 0.142 \\ 
   & (0.0748) & (0.00283) & (0.00285) & (0.00267) & (0.00272) & (0.00265) & (0.00158) & (0.00167) \\ 
  IMDb & 1.02 & 0.954 & 1.05 & 0.922 & 0.883 & \textbf{0.850} & 1.26 & 0.974 \\ 
   & (0.00730) & (0.00721) & (0.00830) & (0.0117) & (0.00837) & (0.00825) & (0.124) & (0.00933) \\ 
  Spotify & 0.0111 & 0.00858 & 0.00880 & \textbf{0.00851} & 0.00894 & 0.00986 & 0.0164 & 0.00926 \\ 
   & (9.20e-05) & (6.23e-05) & (5.79e-05) & (4.69e-05) & (6.99e-05) & (0.000227) & (0.000540) & (8.34e-05) \\ 
  News & 1.88 & 1.89 & 2.37 & 1.85 & 1.78 & \textbf{1.72} & 1.89 & 1.81 \\ 
   & (0.0146) & (0.0147) & (0.0157) & (0.0160) & (0.0143) & (0.0131) & (0.0215) & (0.0190) \\ 
  InstEval & 1.44 & \textbf{1.44} & 1.53 & 1.46 & 1.44 & 1.47 & 1.50 & 1.45 \\ 
   & (0.00672) & (0.00746) & (0.00549) & (0.00711) & (0.00685) & (0.0230) & (0.00669) & (0.00492) \\ 
  Rossmann & 0.0153 & \textbf{0.00627} & 0.0124 & 0.01000 & 0.00910 & 0.00877 & 0.0516 & 0.0105 \\ 
   & (0.000552) & (0.000188) & (0.000848) & (0.000267) & (0.000229) & (0.000284) & (0.00600) & (0.000162) \\ 
  AUimport & 0.743 & 1.26 & 4.56 & 2.15 & 0.714 & \textbf{0.651} & 3.35 & 0.713 \\ 
   & (0.0138) & (0.0269) & (0.0177) & (0.0477) & (0.00550) & (0.00396) & (0.455) & (0.00700) \\ 
  Wages & 0.321 & 0.102 & 0.0992 & 0.0920 & \textbf{0.0829} & 0.0830 &  &  \\ 
   & (0.0222) & (0.00259) & (0.00234) & (0.00251) & (0.00191) & (0.00205) &  &  \\ 
   \hline
Avg. rank & 5.71 & 3.29 & 6.43 & 3.71 & \textbf{2.43} & 2.71 & 7.43 & 4.29 \\ 
  Avg. rel. dif. & 47.7 & 17.4 &    111. & 44.5 & 9.63 & \textbf{8.36} &    189. & 17.3 \\ 
   \hline
\hline
\end{tabular}
\endgroup
\caption{Results when either excluding ('GPBoost\_E') or 
                  including ('GPBoost\_I') the categorical variables 
                  in the fixed effects tree ensemble function for GPBoost.} 
\label{results_high_card_excl_incl_sep}
\end{table}

\end{appendices}

\clearpage
\bibliographystyle{abbrvnat}
\bibliography{bib_Compare_ML_HighCard}

\begin{thebibliography}{18}
\providecommand{\natexlab}[1]{#1}
\providecommand{\url}[1]{\texttt{#1}}
\expandafter\ifx\csname urlstyle\endcsname\relax
  \providecommand{\doi}[1]{doi: #1}\else
  \providecommand{\doi}{doi: \begingroup \urlstyle{rm}\Url}\fi

\bibitem[Avanzi et~al.(2023)Avanzi, Taylor, Wang, and Wong]{avanzi2023machine}
B.~Avanzi, G.~Taylor, M.~Wang, and B.~Wong.
\newblock {Machine Learning with High-Cardinality Categorical Features in
  Actuarial Applications}.
\newblock \emph{arXiv preprint arXiv:2301.12710}, 2023.

\bibitem[Chen and Guestrin(2016)]{chen2016xgboost}
T.~Chen and C.~Guestrin.
\newblock {XGBoost: A scalable tree boosting system}.
\newblock In \emph{Proceedings of the 22nd acm sigkdd international conference
  on knowledge discovery and data mining}, pages 785--794. ACM, 2016.

\bibitem[Fisher(1958)]{fisher1958grouping}
W.~D. Fisher.
\newblock On grouping for maximum homogeneity.
\newblock \emph{Journal of the American statistical Association}, 53\penalty0
  (284):\penalty0 789--798, 1958.

\bibitem[Fu and Simonoff(2015)]{fu2015unbiased}
W.~Fu and J.~S. Simonoff.
\newblock Unbiased regression trees for longitudinal and clustered data.
\newblock \emph{Computational Statistics \& Data Analysis}, 88:\penalty0
  53--74, 2015.

\bibitem[Grinsztajn et~al.(2022)Grinsztajn, Oyallon, and
  Varoquaux]{Grinsztajn2022}
L.~Grinsztajn, E.~Oyallon, and G.~Varoquaux.
\newblock Why do tree-based models still outperform deep learning on typical
  tabular data?
\newblock In S.~Koyejo, S.~Mohamed, A.~Agarwal, D.~Belgrave, K.~Cho, and A.~Oh,
  editors, \emph{Advances in Neural Information Processing Systems}, volume~35,
  pages 507--520. Curran Associates, Inc., 2022.

\bibitem[Guo and Berkhahn(2016)]{guo2016entity}
C.~Guo and F.~Berkhahn.
\newblock Entity embeddings of categorical variables.
\newblock \emph{arXiv preprint arXiv:1604.06737}, 2016.

\bibitem[Hajjem et~al.(2011)Hajjem, Bellavance, and Larocque]{hajjem2011mixed}
A.~Hajjem, F.~Bellavance, and D.~Larocque.
\newblock Mixed effects regression trees for clustered data.
\newblock \emph{Statistics \& probability letters}, 81\penalty0 (4):\penalty0
  451--459, 2011.

\bibitem[Hajjem et~al.(2014)Hajjem, Bellavance, and Larocque]{hajjem2014mixed}
A.~Hajjem, F.~Bellavance, and D.~Larocque.
\newblock Mixed-effects random forest for clustered data.
\newblock \emph{Journal of Statistical Computation and Simulation}, 84\penalty0
  (6):\penalty0 1313--1328, 2014.

\bibitem[Ke et~al.(2017)Ke, Meng, Finley, Wang, Chen, Ma, Ye, and
  Liu]{ke2017lightgbm}
G.~Ke, Q.~Meng, T.~Finley, T.~Wang, W.~Chen, W.~Ma, Q.~Ye, and T.-Y. Liu.
\newblock {LightGBM: A highly efficient gradient boosting decision tree}.
\newblock In \emph{Advances in Neural Information Processing Systems}, pages
  3149--3157, 2017.

\bibitem[Laird et~al.(1982)Laird, Ware, et~al.]{laird1982random}
N.~M. Laird, J.~H. Ware, et~al.
\newblock Random-effects models for longitudinal data.
\newblock \emph{Biometrics}, 38\penalty0 (4):\penalty0 963--974, 1982.

\bibitem[Pinheiro and Bates(2006)]{pinheiro2006mixed}
J.~Pinheiro and D.~Bates.
\newblock \emph{Mixed-effects models in S and S-PLUS}.
\newblock Springer Science \& Business Media, 2006.

\bibitem[Prokhorenkova et~al.(2018)Prokhorenkova, Gusev, Vorobev, Dorogush, and
  Gulin]{CatBoost2017}
L.~Prokhorenkova, G.~Gusev, A.~Vorobev, A.~V. Dorogush, and A.~Gulin.
\newblock {CatBoost: unbiased boosting with categorical features}.
\newblock In \emph{Advances in Neural Information Processing Systems}, pages
  6638--6648, 2018.

\bibitem[Sela and Simonoff(2012)]{sela2012re}
R.~J. Sela and J.~S. Simonoff.
\newblock {RE-EM trees: a data mining approach for longitudinal and clustered
  data}.
\newblock \emph{Machine learning}, 86\penalty0 (2):\penalty0 169--207, 2012.

\bibitem[Shwartz-Ziv and Armon(2022)]{shwartz2022tabular}
R.~Shwartz-Ziv and A.~Armon.
\newblock Tabular data: Deep learning is not all you need.
\newblock \emph{Information Fusion}, 81:\penalty0 84--90, 2022.

\bibitem[Sigrist(2022)]{sigrist2022gaussian}
F.~Sigrist.
\newblock {Gaussian Process Boosting}.
\newblock \emph{The Journal of Machine Learning Research}, 23\penalty0
  (1):\penalty0 10565--10610, 2022.

\bibitem[Sigrist(2023)]{sigrist2023gaussian}
F.~Sigrist.
\newblock {Latent Gaussian Model Boosting}.
\newblock \emph{IEEE Transactions on Pattern Analysis and Machine
  Intelligence}, 45\penalty0 (2):\penalty0 1894--1905, 2023.

\bibitem[Simchoni and Rosset(2021)]{simchoni2021using}
G.~Simchoni and S.~Rosset.
\newblock {Using random effects to account for high-cardinality categorical
  features and repeated measures in deep neural networks}.
\newblock \emph{Advances in Neural Information Processing Systems},
  34:\penalty0 25111--25122, 2021.

\bibitem[Simchoni and Rosset(2023)]{simchoni2023integrating}
G.~Simchoni and S.~Rosset.
\newblock {Integrating Random Effects in Deep Neural Networks}.
\newblock \emph{Journal of Machine Learning Research}, 24\penalty0
  (156):\penalty0 1--57, 2023.

\end{thebibliography}

\end{document}